\documentclass[conference]{IEEEtran}
\IEEEoverridecommandlockouts
\usepackage{cite}
\usepackage{amsmath,amssymb,amsfonts}
\usepackage{algorithmic}
\usepackage{graphicx}
\usepackage{textcomp}
\usepackage{xcolor}
\def\BibTeX{{\rm B\kern-.05em{\sc i\kern-.025em b}\kern-.08em
		T\kern-.1667em\lower.7ex\hbox{E}\kern-.125emX}}

\usepackage[switch]{lineno} 
\usepackage{hyperref}
\usepackage{siunitx}
\usepackage{booktabs} 
\usepackage{amssymb}
\usepackage{amsthm,amsmath,amssymb}
\usepackage{mathrsfs}
\usepackage{array}
\usepackage{multicol}
\usepackage{multirow}
\usepackage{stfloats}
\usepackage{threeparttable}
\graphicspath{{figures/}}
\usepackage{algorithm}
\usepackage{algorithmic}

\begin{document}
	
	\title{ObjectAug: Object-level Data Augmentation for Semantic Image Segmentation
	}
	
	\author{
		\IEEEauthorblockN{Jiawei Zhang\IEEEauthorrefmark{1}\IEEEauthorrefmark{2}, Yanchun Zhang\IEEEauthorrefmark{2}\IEEEauthorrefmark{3}, Xiaowei Xu\IEEEauthorrefmark{4}}
		\IEEEauthorblockA{\IEEEauthorrefmark{1}Shanghai key Laboratory of Data Science, School of Computer Science, Fudan University, Shanghai, China\\
		}
		\IEEEauthorblockA{\IEEEauthorrefmark{2}Cyberspace Institute of Advanced Technology, Guangzhou University, Guangzhou, China\\
		}
		\IEEEauthorblockA{\IEEEauthorrefmark{3}College of Engineering and Science, Victoria University, Melbourne, Australia\\
		}
		\IEEEauthorblockA{\IEEEauthorrefmark{4}
			Guangdong Cardiovascular Institute, Guangdong Provincial People's Hospital, Guangzhou, China\\
		}
		Email: 17110240008@fudan.edu.cn, yanchun.zhang@vu.edu.au, xiao.wei.xu@foxmail.com
	}
	
	\maketitle
	
	\begin{abstract}
		Effective training of deep neural networks (DNNs) usually requires labeling a large dataset, which is time and labor intensive.
		Recently, various data augmentation strategies like regional dropout and mix strategies have been proposed, which are effective as the augmented dataset can guide the model to attend on less discriminative parts.
		However, these strategies operate only at the image level, where the objects and the background are coupled.
		Thus, the boundaries are not well augmented due to the fixed semantic scenario. 
		In this paper, we propose ObjectAug to perform object-level augmentation for semantic image segmentation.
		Our method first decouples the image into individual objects and the background using semantic labels.
		Second, each object is augmented individually with commonly used augmentation methods (e.g., scaling, shifting, and rotation). 
		Third, the pixel artifacts brought by object augmentation are further restored using image inpainting.
		Finally, the augmented objects and background are assembled as an augmented image.
		In this way, the boundaries can be fully explored in the various semantic scenarios. 
		In addition, ObjectAug can support category-aware augmentation that gives various possibilities to objects in each category, and can be easily combined with existing image-level augmentation methods to further boost the performance.
		Comprehensive experiments are conducted on both natural image and medical image datasets.
		Experiment results demonstrate that our ObjectAug can effectively improve segmentation performance.
	\end{abstract}
	
	\begin{IEEEkeywords}
		Object-level, Data Augmentation, Image Segmentation
	\end{IEEEkeywords}

	\begin{figure}[t]
		\centering
		\includegraphics[width=0.85\linewidth]{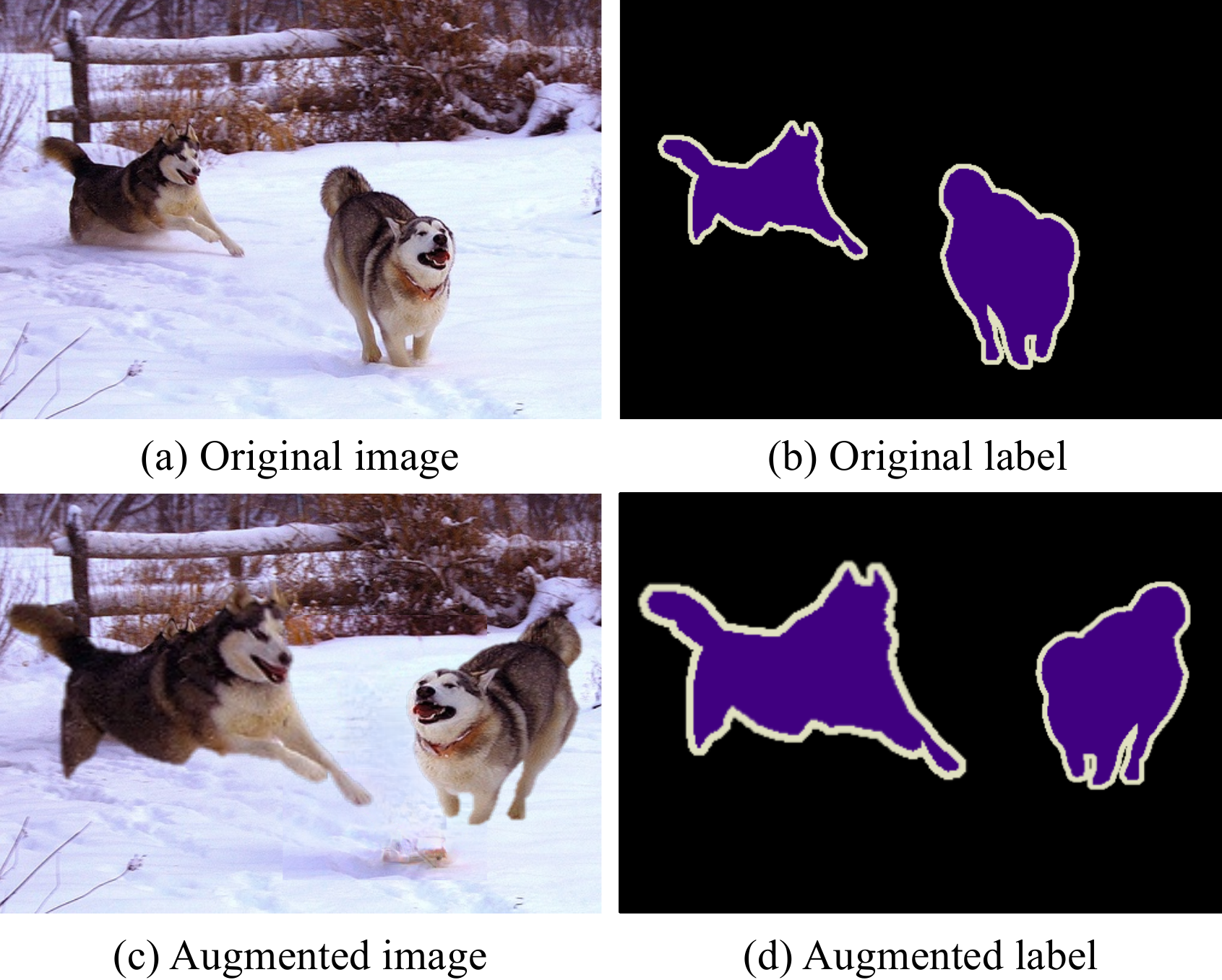}
		\caption{
			Illustration of ObjectAug. ObjectAug operates at the object level, and can perform various augmentation methods for each object as shown in (c) and (d). The left husky is scaled and shifted, while the right one is flipped and shifted.
			Thus, the boundaries are extensively augmented to boost the performance of semantic segmentation.
		}
		\label{fig-traditional}
	\end{figure}

	\begin{figure*}[t]
		\centering
		\includegraphics[width=0.97\linewidth]{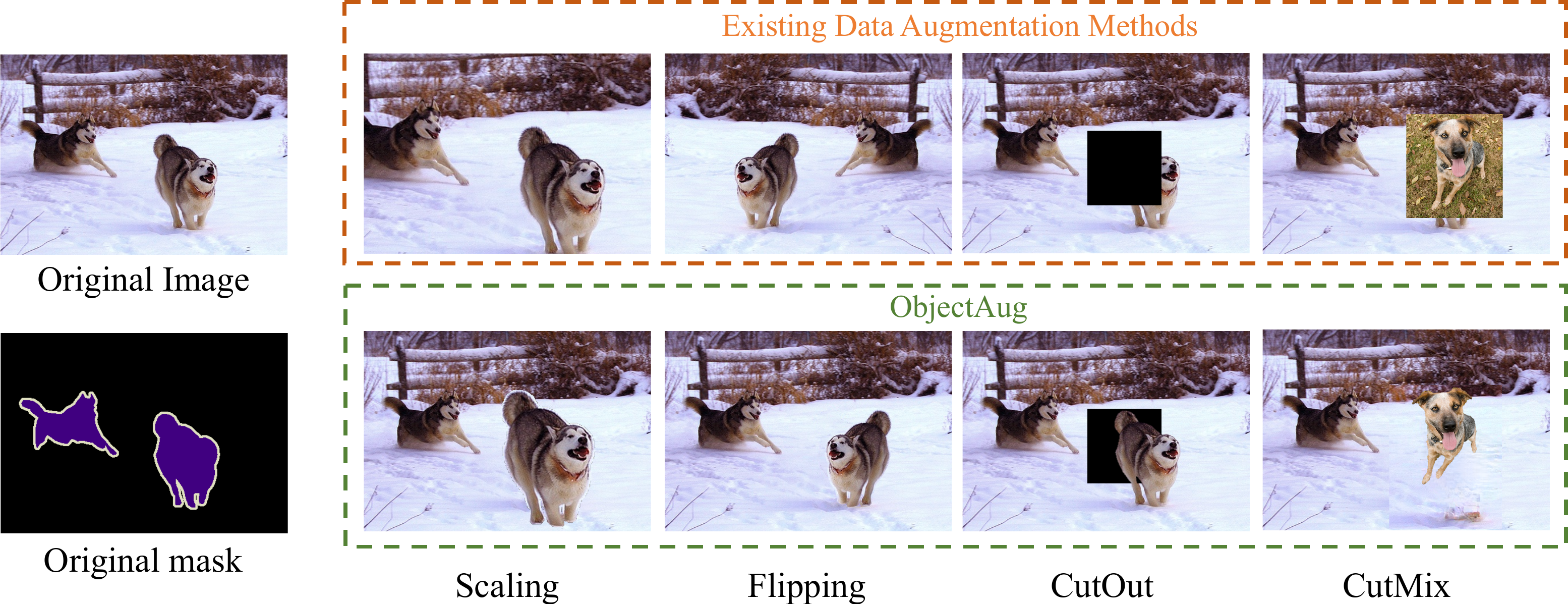}
		\caption{
			Comparison of ObjectAug and existing data augmentation methods.
			All the existing methods operate at the image level, while ObjectAug at the object level.
			ObjectAug can adopt existing methods to augment each object in the image.
		}
		\label{fig-compare}
	\end{figure*}

	\section{Introduction}

	Semantic segmentation with the goal to assign semantic labels to target objects in an image is one of the fundamental topics in computer vision \cite{xu2018quantization,xu2018scaling}. 
	Recently, deep neural networks (DNNs) have been widely adopted in semantic segmentation with boosted performance
	\cite{fcn,resnet,unet,mdunet,vgg,previous,new2,pspnet,deeplabv3,deeplabv3p}.
	Typical DNN requires a large amount of labeled data for training \cite{imagenet,cityscape}, which is a time and labour intensive process.
	%
	To get around this, machine learning practitioners typically either transfer knowledge from other tasks (i.e. transfer learning \cite{transfer}) or generate synthetic data based on labeled data (i.e. data augmentation).
	%
	Various classical data augmentation strategies, including random rotation, random scaling and random cropping \cite{vgg,class2} artificially expanded the size of datasets by orders of magnitude, thus with improved performance in most vision tasks. 

	Recently, several works \cite{cutout,mixup,cutmix,randomerase,augmix,mmixup,comixup,autoaugment} 
	proposed to mix data at the image level (a.k.a mixed images) to augment the data which can guide the model to attend on less discriminative parts.
	%
	For example, CutOut \cite{cutout} and random erase \cite{randomerase} randomly masked out square regions of the input during training, which improves the robustness and overall performance of convolutional neural networks.
	CutMix \cite{cutmix} generated a new training sample by randomly combining two cropped training samples, and achieves a better performance than CutOut.
	However, both methods operate at the image level, where the objects and the background are coupled in the image.
	This means that the boundaries between objects and the background are fixed, which are important to obtain precise segmentation \cite{ding2019boundary,kervadec2019boundary}.
	%
	%
	%
	In this way, the critical boundaries are not augmented well.
	It occurs to us: is it possible to effectively augment the boundaries in semantic segmentation to further boost the performance?

	In this paper, we propose an object-level augmentation method, ObjectAug, to address the above question. 
	As shown in Figure \ref{fig-traditional}, instead of dealing with the image directly, ObjectAug utilizes the segmentation annotations to decouple the image into the background and objects.
	The object-removed background can be further restored using image inpainting, while the objects can be augmented independently.
	%
	The annotations are also processed in a similar way as that of their corresponding images. 
	%
	%
	Finally, the augmented objects and the background are assembled as an augmented image in which objects can be placed flexibly on the background.
	%
	During the augmentation, ObjectAug usually misaligns areas of the original object and the enhanced object, which would cause pixel artifacts (the area that belongs to the original object and not to the enhanced object).
	%
	%
	Since the augmentations for each object are performed independently, category-aware augmentation can be easily achieved, e.g., objects in some categories are augmented in every augmented image, while others remain (unaugmented) in some augmented images.
	%
	On the other hand, our ObjectAug breaks the consistency among the objects and diversifies their compositions, the robustness is therefore further improved.
	The performance can be further boosted by combining the existing augmentation methods since they are on image level, which works at a higher level than that of our objectAug (object level).
	%
	%
	We present extensive evaluations of ObjectAug on various DNN architectures and semantic image segmentation datasets.
	Experiments across various architectures and datasets demonstrate that our ObjectAug can effectively improve the segmentation performance.

	In summary, our contributions follow,
	\begin{itemize}
		\item We propose an object-level data augmentation method, ObjectAug, which can be easily combined with other existing image-level augmentation methods to further boost performance.  
		%
		%
		\item ObjectAug can perform category-aware augmentation to mitigate the category imbalance problems,
		which performs diversified augmentation methods to objects belonging to different categories.
		\item Extensive experiments across various networks and datasets demonstrate that ObjectAug 
		outperforms existing data augmentation methods and significantly improves the performance of segmentation.
		
	\end{itemize}

	\section{Related work}
	
	Existing augmentation methods can be divided into two general approaches: traditional augmentation methods, and DNN-orientated augmentation methods. 
	%
	
	The traditional augmentation approach has been widely used including random crops, horizontal flipping, and color augmentation \cite{class1}, which could improve the robustness of translated, reflected, and illuminated objects, respectively. 
	Random scaling \cite{vgg} as well as random rotations and affine transformations are also widely adopted.
	Geometric distortions or deformations are commonly used to increase the number of samples for training DNNs to balance datasets.
	The above methods have proven to be fast, reproducible, and reliable, and its implementation code is relatively easy and available for the most known deep learning frameworks, which makes it even more widespread \cite{effectiveness}.

	The DNN-orientated augmentation approach is an emerging trend, which takes the learning characteristics of DNNs into consideration.
	Mixup \cite{mixup} uses information from two images. Rather than implanting one portion of an image inside another, Mixup produces an element-wise convex combination of two images. 
	An adaptive mixing policy \cite{mixup1} was proposed to improve Mixup preventing manifold intrusion. 
	CutOut \cite{cutout} randomly masks out square regions of the input during training, which improves the robustness and overall performance.
	Rather than occluding a portion of an image, CutMix \cite{cutmix} generates a new training sample by randomly combining two cropped training samples.
	Moreover, Generative Adversarial Network (GAN) \cite{gan} is getting popular in data augmentation domain. 
	For example, \cite{pixel} employed a GAN to generate realistic images by reconstructing the image, while our method aims on augmenting the true objects in the image rather than generating the new objects.
	The methods are out of the scope of this paper.
	%
	
	
	%
	In this paper, we compared ObjectAug with CutOut and CutMix, which are the most popular approaches in DNN-oriented augmentation method.
	Note that CutOut and CutMix are primarily for classification and detection tasks, which can be easily revised for segmentation tasks (detailed in the experiment section).
	Illustration of comparison between ObjectAug, traditional data augmentation methods (scaling and flipping), CutOut and CutMix are shown in Figure \ref{fig-compare}.

	\begin{figure*}[ht]
		\centering
		\includegraphics[width=0.99\linewidth]{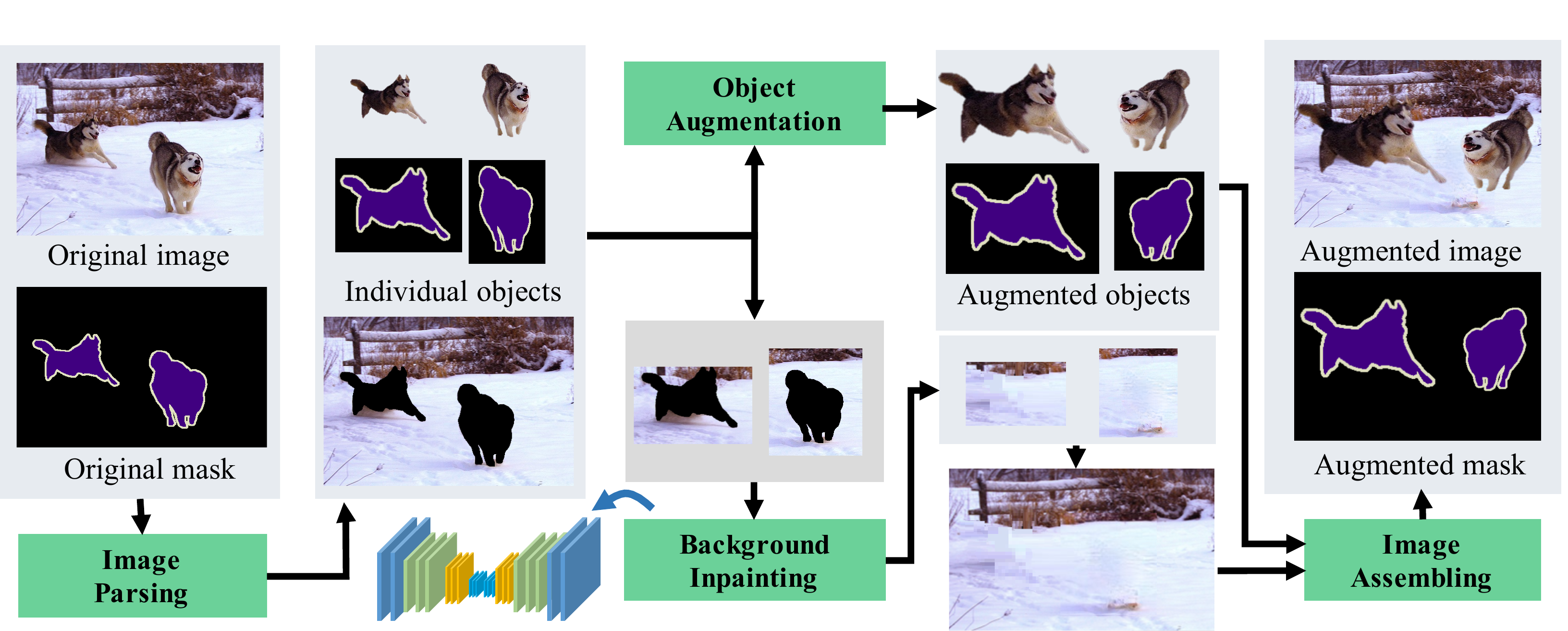}
		\caption{
			Overall flow of ObjectAug.
			ObjectAug includes four modules: image parsing, object augmentation, background inpainting, and image assembling. 
			First, image parsing decouples the image and extracts multiple objects in the image. 
			Next, object augmentation adopts various data augmentation methods to augment the objects.
			%
			At the same time, background inpainting restores the object-removed background to deal with the pixel artifacts.
			Finally, the augmented objects with the restored background and their masks are assembled to obtain the augmented image and mask.
		}
		\label{fig-overall}
	\end{figure*}

	\section{Method}

	%
	\subsection{Overview}
	The overall flow of ObjectAug is shown in Figure \ref{fig-overall}, and Algorithm \ref{alg:Framwork} details the process of ObjectAug. 
	ObjectAug includes four modules: image parsing, object augmentation, background inpainting, and image assembling. 
	First, image parsing decouples the image and extracts the multiple objects in the image. 
	Next, the object augmentation performs various data augmentation methods to augments the objects.
	%
	At the same time, the background inpainting restores the object-removed background to deal with the pixel artifacts.
	Finally, the augmented objects, the corresponding masks, and the restored background are assembled by the image assembling module as the augmented image and mask.
	We detail the four modules of ObjectAug as follows.

	\subsection{Image Parsing}
	Image parsing decouples the image and extracts the objects from the image.
	For ease of discussion, a training image and its mask are denoted as $ \mathbf{I}$ and $\mathbf{M}$, respectively.
	By utilizing the segmentation annotation, we can easily decouple the mask $\mathbf{M}$ into $ \mathbf{M}^{k} \in\{0,1\}^{W \times H} $, which is the mask of the $k$-th object.
	It associates binary values in the mask to pixels in the image so that $\mathbf{M}^k_{x,y} = 1$ if the pixel at $(x, y)$ belongs to the $k$-th object. 
	Note that since one pixel can corresponds to only one object, the masks are constrained by Eq. (\ref{con:E1}).
	$\mathbf{1}$ presents the all-ones matrix.
	
	\begin{equation}
	\begin{aligned}
	\sum_{k=1}^{n} \mathbf{M}^{k}= \mathbf{1},  \quad
	\mathbf{M}^{n}=\mathbf{1}-\sum_{k=1}^{n-1} \mathbf{M}^{k}.
	\label{con:E1}
	\end{aligned}
	\end{equation}

	The background masks $\mathbf{M}_n$ can be easily retrieved from the object masks.
	%
	The process of extracting the $k$-th object is denoted as
	$
	\mathbf{I}^k \in \mathbb{R}^{W \times H \times C}.
	$
	The decouple of the objects in an image is computed as follows,
	
	\begin{equation}
	\begin{aligned}
	\mathbf{I}^{k} = \mathbf{M}^{k} \odot \mathbf{I}, \quad
	\sum_{k=1}^{n}  \mathbf{I}^{k} = \mathbf{I}.
	\label{con:E2}
	\end{aligned}
	\end{equation}
	where $\odot$ is dot product.
	Meanwhile, if the traditional data augmentation methods such as rotation or flipping are used directly to the object with its size unchanged, its position may change considerably, which may result in pixel artifacts in the image.
	In order to minimize the pixel artifacts brought by ObjectAug, we crop each object $\mathbf{I}^k$ and its corresponding ground truth $\mathbf{M}^k$, and obtain the cropped object patch $\mathbf{I}^k_{c}$ and the ground truth $\mathbf{M}^k_{c}$.
	$\mathbf{\psi }$ denotes the crop parameter, which takes the center of the object as the cropping center, and the crop size is based on the size of the object.
	%
	%
	
	\begin{equation}
	\begin{aligned}
	\mathbf{I}^{k}_{c}, \mathbf{M}^{k}_{c} = f_{c}( \mathbf{I}^{k}, \mathbf{M}^{k} | \mathbb{\psi} ).
	\end{aligned}
	\label{con:E3}
	\end{equation}

	\subsection{Object Augmentation}
	The object augmentation module augments the extracted objects from the image parsing module individually.
	Given a series of traditional data augmentation methods including scaling, rotation, shifting, flipping, brightening and etc., denoted as [$f_1, f_2, ..., f_m$], they are performed to each object with a set of corresponding probabilities $\mathbf{P}$=[$p_1, p_2, ..., p_m$].
	With the sequentially performing the augmentations, we obtain the composed augmentation function ${f}_{ObjAug}$.
	We then apply ${f}_{ObjAug}$ to the extracted object and mask patches, and obtain the augmented objects.
	
	\begin{equation}
	\begin{aligned}
	{f}_{ObjAug}(\mathbf{I} | \mathbf{P}) = {f}_1(\mathbf{I} | p_1) \circ f_2(\mathbf{I}|p_2) \circ \ldots \circ f_m(\mathbf{I}|p_m).
	\end{aligned}
	\label{con:E4}
	\end{equation}

	\begin{equation}
	\begin{aligned}
	\mathbf{I}^{k}_{aug}, 
	\mathbf{M}^{k}_{aug} = f_{ObjAug}( \mathbf{I}^{k}_{c},  \mathbf{M}^{k}_{c} | \mathbf{P} ).
	\end{aligned}
	\label{con:E5}
	\end{equation}
	
	As each object can be augmented individually, the object augmentation module supports category-aware augmentation.
	Note that the category imbalance is a common issue in semantic segmentation, where the categories with the less and difficult objects to segment are more critical in training.
	%
	%
	%
	Meanwhile, the data augmentation not only brings generalization to the train samples, it also makes the training more difficult.
	Therefore, an intuitive idea is that the method can be inclined to more rare cases called rarity-driven coefficient so that the model can alleviate under-fitting because of too few examples.
	However, we actually find that the rarity of objects is not directly proportional to the final segmentation performance.
	For example, the number of object in category people is far more than that of all other category objects, but its segmentation performance is ranked 13-th.
	Then we also use the segmentation performance as the criterion of the category-aware coefficient called hard-driven coefficient.
	We will detail the comparison of two coefficients in ablation analysis.
	%
	For the hard-driven coefficient $\alpha$, $p_j$ represents the performance in the previous experiment of the $j$-th category, and $\hat p$ represents the median of the performance.
	In rarity-driven coefficients $\beta$, 
	$\hat N$ represents the median of the number of objects and $N_j$ represents the number of objects belonging to the $j$-th category.
	$\alpha_j$/$\beta_j$ is defined as the hard-driven/rarity-driven coefficient for the $j$-th category as follows,
	
	\begin{equation}
	\begin{aligned}
	\alpha_j = \frac{\hat p}{p_j},\beta_j = \frac{\hat N}{N_j},
	\end{aligned}
	\label{con:E6}
	\end{equation}
	Finally, the probability of the $j$-th category using the augmentation method $m$ and the hard-driven coefficient $\alpha$ is defined as,
	
	\begin{equation}
	\begin{aligned}
	p_{j,m} =  p_{m} \times {\alpha}_j.
	\end{aligned}
	\label{con:E7}
	\end{equation}

	\begin{algorithm}[htb] 
		\caption{ Process of ObjectAug.} 
		\label{alg:Framwork} 
		\begin{algorithmic}[1] 
			\REQUIRE ~~\\ 
			An original image $\mathbf{I}$ and its annotation $\mathbf{M}$ containing $n$ objects;
			Selected data augmentation methods [$f_1, f_2, ..., f_m$] 
			with probabilities [$p_1, p_2, ..., p_m$],
			category-aware coefficient [${\alpha}_1,{\alpha}_2, ..., {\alpha}_J$], $J$ is the number of class; \\
			\ENSURE ~~\\ 
			Augmented image $\mathbf{I}$ and its annotation $\mathbf{M}$;\\
			
			\STATE 
			Parse
			$
			\sum_{k=1}^{n}  \mathbf{I}^{k} = \mathbf{I}
			$
			\hspace{2.8cm}  \# \textbf{Parsing Stage} \\  
			\STATE 
			Parse
			$
			\sum_{k=1}^{n}  \mathbf{M}^{k} = \mathbf{1}
			$

			\STATE 
			$
			\mathbf{I}^{k}_{c}, \mathbf{M}^{k}_{c}  \leftarrow  f_{c}( \mathbf{I}^{k}, \mathbf{M}^{k}, \psi  )
			$

			\STATE \textbf{for} all objects $I^k$  \textbf{do} \hspace{1.5cm} \hspace{0.3cm} \# \textbf{Augmenting Stage}

			\STATE  \quad 
			$
			\mathbf{I}^{k}_{aug}, \mathbf{M}^{k}_{aug} \leftarrow  f_{ObjectAug}( \mathbf{I}^{k}_{c}, 
			\mathbf{M}^{k}_{c} , p_{j,m}  )
			$
			\STATE  \quad 
			$
			\mathbf{I}_i^{k}\leftarrow \Phi(\mathbf{I}_{c}, \mathbf{M}_{c}^{k})
			$
			\hspace{2.4cm} \# \textbf{Inpainting Stage}

			\STATE  \quad 
			$
			\mathbf{I}_{c}, \mathbf{M}_{c}  \leftarrow  f_{c}( \mathbf{I}, \mathbf{M}, \psi  )
			$ 
			\hspace{1.5cm} \# \textbf{Assembling Stage} \\

			\STATE \quad 
			$
			\widehat{\mathbf{M}}_{c}^{k} \leftarrow  \mathbf{M}^k_{c} \cup \mathbf{M}^k_{aug}, \quad
			\widehat{\mathbf{M}}_{c}^{k'} \leftarrow  \widehat{\mathbf{M}}_{c} - \mathbf{M}^k_{aug} 
			$

			\STATE \quad 
			$
			\mathbf{I}_{asm}  \leftarrow   \mathbf{I}_{c} \odot (\mathbf{1}- \widehat{\mathbf{M}}_{c}) + \mathbf{I}^k_{aug} + \mathbf{I}_i^{k} \odot \widehat{\mathbf{M}}_{c}^{''}
			$
			\STATE \quad 
			$
			\mathbf{M}_{asm} \leftarrow   \mathbf{M}_{c} \odot (\mathbf{1}- \widehat{\mathbf{M}}_{c}) + \mathbf{M}^k_{aug}.
			$
			

		\end{algorithmic}
	\end{algorithm}

	\subsection{Background Inpainting}
	The image parsing module brings pixel artifacts in the object-removed background which can not always be covered completely by the augmented object $I^k_{aug}$.  
	%
	%
	In addition to filling random noise,
	we use image inpainting methods to fill the pixel artifacts. 
	%
	%
	To prevent the inpainting methods from being affected by objects, we inpaint the background with the object removed.
	Particularly, we perform morphological processing to expand extracted object masks thus to remove residual edges.
	We employ the inpainting methods from \cite{inpaint1} and denote it as $\Phi$, which uses DNNs with partial convolutions to deal with the irregular holes.
	The DNN models for image inpainting is trained on ImageNet \cite{iimagenet}.
	The background inpainting can be described as,
	\begin{equation}
	\begin{aligned}
	\mathbf{I}_i^{k}=\Phi(\mathbf{I}_{c}, \mathbf{M}_{c}^{k}).
	\end{aligned}
	\label{con:E8}
	\end{equation}
	where $\mathbf{I}_i^{k}$ is the inpainted result from the original cropped image $\mathbf{I}_{c}$ with selected object $\mathbf{M}_{c}^{k}$ removed.
	In the inpainting process, the images are all scaled to a uniform size.
	Meanwhile, since the object is removed and we only inpaint the background, the annotation does not need to be processed accordingly.


	\begin{figure}[ht]
		\centering
		\includegraphics[width=1\linewidth]{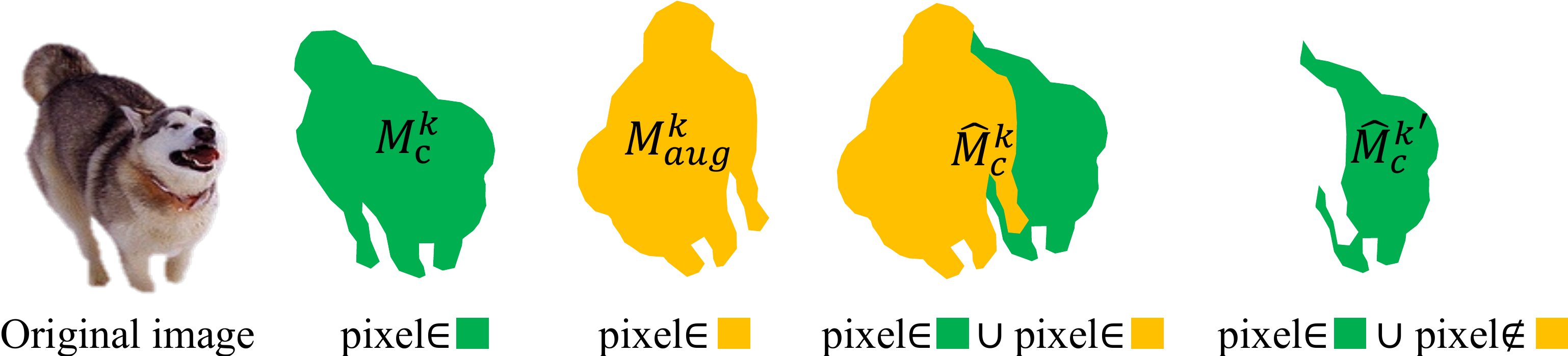}
		\caption{
			Visualization of masks $\widehat{\mathbf{M}}_{c}^{k}$ and $\widehat{\mathbf{M}}_{c}^{k'}$ in the image assembling step.
		}
		\label{fig-mis}
	\end{figure}

	\subsection{Image Assembling}
	%
	In image assembling, the augmented objects and the restored background are combined into the augmented images and masks.
	First, we use the identical cropping parameter $\psi$ to directly crop the original image $\mathbf{I}$ and annotation $\mathbf{M}$ to obtain the cropped image $\mathbf{I}_{c}$ and annotation $\mathbf{M}_{c}$.
	
	\begin{equation}
	\begin{aligned}
	\mathbf{I}_{c}, \mathbf{M}_{c} = f_{c}( \mathbf{I}, \mathbf{M} | \psi ).
	\end{aligned}
	\label{con:E9}
	\end{equation}

	%
	Next, the masks of the augmented objects are processed following,
	\begin{equation}
	\begin{aligned}
	\widehat{\mathbf{M}}_{c}^{k} = \mathbf{M}^k_{c} \cup \mathbf{M}^k_{aug}, \quad
	\widehat{\mathbf{M}}_{c}^{k'} = \widehat{\mathbf{M}}_{c} - \mathbf{M}^k_{aug},
	\end{aligned}
	\label{con:E10}
	\end{equation}
	where $\widehat{\mathbf{M}}^{k}_{c}$ denotes the union of the original annotation and the augmented annotation for the current image patch $\mathbf{I}_{c}$; 
	$\widehat{\mathbf{M}}_{c}^{k'}$ is the pixel artifacts introduced by the misaligned original object and augmented object.
	For image assembling, the corresponding image area of $\widehat{\mathbf{M}}^{k}_{c}$ is reset (to be pixel artifacts), while the corresponding image area of $\widehat{\mathbf{M}}_{c}^{k'}$ is restored using image inpainting.
	%

	%
	%
	Finally, we assemble the augmented object $I^k_{aug} $ into the cropped original image as Eq. (\ref{con:E11}).
	Meanwhile, the pixel artifacts is filled with the inpainting results $\mathbf{I}_i^{k}$. 
	
	\begin{equation}
	\begin{aligned}
	\mathbf{I}_{asm}^{'}  =  \mathbf{I}_{c} \odot (\mathbf{1}- \widehat{\mathbf{M}}^{k}_{c}) + I^k_{aug} + \mathbf{I}_i^{k} \odot \widehat{\mathbf{M}}_{c}^{k'}.
	\end{aligned}
	\label{con:E11}
	\end{equation}
	
	\begin{equation}
	\begin{aligned}
	\mathbf{M}_{asm} =  \mathbf{M}_{c} \odot (\mathbf{1}-\widehat{\mathbf{M}}_{c}^{k}) + \mathbf{M}^k_{aug}.
	\end{aligned}
	\label{con:E12}
	\end{equation}

	By assembling the assembled patches $\mathbf{I}_{asm}$ into the image with the relevant area removed, we get the image with the $k$-th object augmented.
	Similarly, the corresponding object-level augmented ground truth is obtained accordingly. 
	We then repeat the above process until all the objects are processed, and the augmented image and annotation are obtained.

	\section{Experiments}
	In this section, ObjectAug is extensively evaluated and compared with the existing methods on public datasets. 
	%
	We first describe our experimental setup in detail.
	Then we compare ObjectAug with CutOut and CutMix, and proceed ablation analysis of hyper-parameters, image inpainting, and category-aware coefficient on the PASCAL VOC 2012 segmentation dataset. 
	At last, we conduct extended experiments on the Cityscapes dataset and CRAG dataset to show the generalizability of our method. 

	\subsection{Experiment Setup}
	
	\textbf{Datasets}
	Three image segmentation datasets including PASCAL VOC 2012, Cityscapes, and CRAG are used for evaluation.
	
	\begin{itemize}
		\item \textbf{PASCAL VOC 2012.}
		The PASCAL VOC 2012 semantic
		segmentation benchmark \cite{voc} contains 20 foreground object classes.
		%
		The original dataset contains 1, 464 (train), 1, 449 (val), and 1, 456 (test) pixel-level annotated images.
		%
		%
		The dataset is augmented by the extra annotations provided by \cite{trainaug}, resulting in 10, 582 (trainaug) training images;
		
		\item \textbf{Cityscapes.}
		The Cityscapes dataset \cite{cityscape} is a large-scale dataset containing high-quality pixel-level annotations of 5000 images (2975, 500, and 1525 for training, validation, and test, respectively) and about 20000 coarsely annotated images;
		
		\item \textbf{CRAG.}
		The Colorectal Adenocarcinoma Gland (CRAG) dataset \cite{mild} has a total of 213 H\&E CRA images taken from 38 WSIs.
		%
		Images are at 20$\times$ magnification and are mostly of size 1512$\times$1516 pixels, with corresponding instance-level ground truth.
		The CRAG dataset is split into 173 training images and 40 test images with different cancer grades. 

	\end{itemize}

	\noindent
	\textbf{Implementation Details} 
	All experiments were implemented and evaluated on PyTorch platform with Titan X (Pascal) with 12 GB memory. 
	For the PASCAL VOC 2012 and cityscapes datasets, we followed the same training settings as in \cite{deeplabv3}.
	%
	In short, we employed the same learning rate schedule (i.e., “poly” policy \cite{parsenet}  and same initial learning rate 0.007), cropped size 513 $\times$ 513, fine-tuning batch normalization parameters with output stride 16, and random scale data augmentation during training. 
	Besides, we also used some traditional data augmentation methods including random scale, random shift, random rotation, and random horizontal flip to augment objects in ObjectAug.
	The specific setting of ObjectAug details in the result and discussion section. 
	The performance was measured in terms of pixel intersection-over-union averaged across all classes (mIOU).
	For image inpainting, we utilized ResNet50 based U-Net as our model and the loss function from \cite{inpaint1}.
	We implemented the model with the pre-trained model on ImageNet \cite{imagenet} and finetuned it on current datasets to enhance the inpainting performance.

	\begin{table}[ht]
		\centering
		\caption{
			Performance comparison of models with and without ObjectAug on PASCAL VOC 2012.
		}
		\begin{tabular}{p{2.3cm}<{\centering}||p{1.6cm}<{\centering}||p{1.5cm}<{\centering}||p{1.5cm}<{\centering}}
			\toprule[2pt]
			Method      &Model &ObjectAug & mIoU (\%) \\  \hline		
			
			\multirow{6}*{DeepLab V3}&\multirow{2}*{MobileNet } & $\times$ &70.1 \\
			&        & \checkmark &\textbf{71.9 }\\ \cline{2-4}
			
			
			&\multirow{2}*{ResNet-50 }      & $\times$ &76.9 \\
			&     & \checkmark &\textbf{77.8} \\ \cline{2-4}
			
			&\multirow{2}*{ResNet-101 }      & $\times$  &77.3 \\
			&      & \checkmark &\textbf{78.4} \\ \hline
			
			\multirow{6}*{DeepLab V3plus}&\multirow{2}*{MobileNet }       & $\times$ &71.4 \\
			&        & \checkmark &\textbf{73.8}  \\ \cline{2-4}
			
			
			&\multirow{2}*{ResNet-50 }      & $\times$ &77.2 \\
			&     & \checkmark &\textbf{78.8}  \\ \cline{2-4}
			
			&\multirow{2}*{ResNet-101 }      & $\times$ &78.3 \\
			&      & \checkmark &\textbf{79.6} \\ 
			\bottomrule[2pt]
		\end{tabular}
		\label{tab-voc}
	\end{table}
	
	\begin{table}[t]
		\begin{threeparttable}
			\centering
			\caption{
				Performance comparison of ObjectAug and traditional data augmentation methods on PASCAL VOC 2012. 
			}
			\begin{tabular}{p{2.3cm}<{\centering}||p{1.6cm}<{\centering}||p{1.5cm}<{\centering}||p{1.5cm}<{\centering}}
				\toprule[2pt]
				
				Method &Image level& Object level& mIoU (\%)\\ \hline	
				
				Baseline   &&&68.5        \\ \hline	
				
				\multirow{3}*{+ R.Rotation}   &\checkmark&&69.8     \\ 
				&&\checkmark&   69.5   \\  \hline	
				
				\multirow{3}*{+ R.Scaling}   &\checkmark&&69.9       \\ 
				&&\checkmark&70.3      \\ \hline	
				
				\multirow{3}*{+ R.Flipping}   &\checkmark&&70.1      \\ 
				&&\checkmark& 69.6      \\ \hline	
				
				\multirow{3}*{+ R.Shifting}   &\checkmark&&70.3     \\ 
				&&\checkmark&70.7        \\ \hline	
				
				Baseline + All   &\checkmark&&71.4       \\ 
				Baseline + All   &&\checkmark&71.2        \\ 
				Baseline + All    &\checkmark&\checkmark&73.8       \\ 
				\bottomrule[2pt]	
			\end{tabular}
			
			\begin{tablenotes}
				\item[*]  All means the combination of the above four data augmentation methods.
			\end{tablenotes}
			\label{tab-image-level}
		\end{threeparttable}
		
	\end{table}

	\subsection{Results on PASCAL VOC 2012}
	
	\subsubsection{Effectiveness of ObjectAug}
	We evaluated ObjectAug based on DeepLab V3 \cite{deeplabv3} and DeepLab V3plus \cite{deeplabv3p} with three backbone model MobileNet \cite{mobilenetv2}, ResNet-50 \cite{resnet}, and ResNet-101 on the PASCAL VOC 2012 dataset.
	Note that all baseline are implemented with traditional data augmentation methods including random scaling, random rotation and random flipping.
	%
	As shown in Table \ref{tab-voc}, we can see that ObjectAug improves the mIoU of MobileNet, ResNet-50 and ResNet-101 based DeepLab V3plus by 1.8\%, 0.9\%, and 1.1\%, respectively.
	Meanwhile, MobileNet, ResNet-50 and ResNet-101 based DeepLab V3 obtain an improvement of 2.4\%, 1.6\%, and 1.3\%, respectively.
	
	%


	\subsubsection{Comparison with Traditional Data Augmentation Methods}
	%
	In this part, we select four widely-used traditional data augmentation methods including random rotation, random shifting, random scaling, and random flipping for comparison. 
	Note that the above operations can be performed at two levels: image level (the traditional approach) and object-level in ObjectAug.
	%
	As shown in Table \ref{tab-image-level}, the operation at the object level achieves competitive performance with the methods at the image level.
	Surprisingly, ObjectAug with random scaling at the object level outperforms the methods at the image level slightly.
	On the other hand, we also evaluated the combination of object-level augmentation and image-level augmentation.
	Notably, we can observe that the combination approach achieves much better results than each of them, which indicates that ObjectAug has no conflict with traditional data augmentation methods.
	This is due to the fact that the two kinds of augmentation methods operate at different levels, and thus they can exploit semantic information at different scales without overlap.  
	%

	\subsubsection{Comparison with DNN-based Methods}
\begin{figure}[t]
		\centering
		\includegraphics[width=0.79\linewidth]{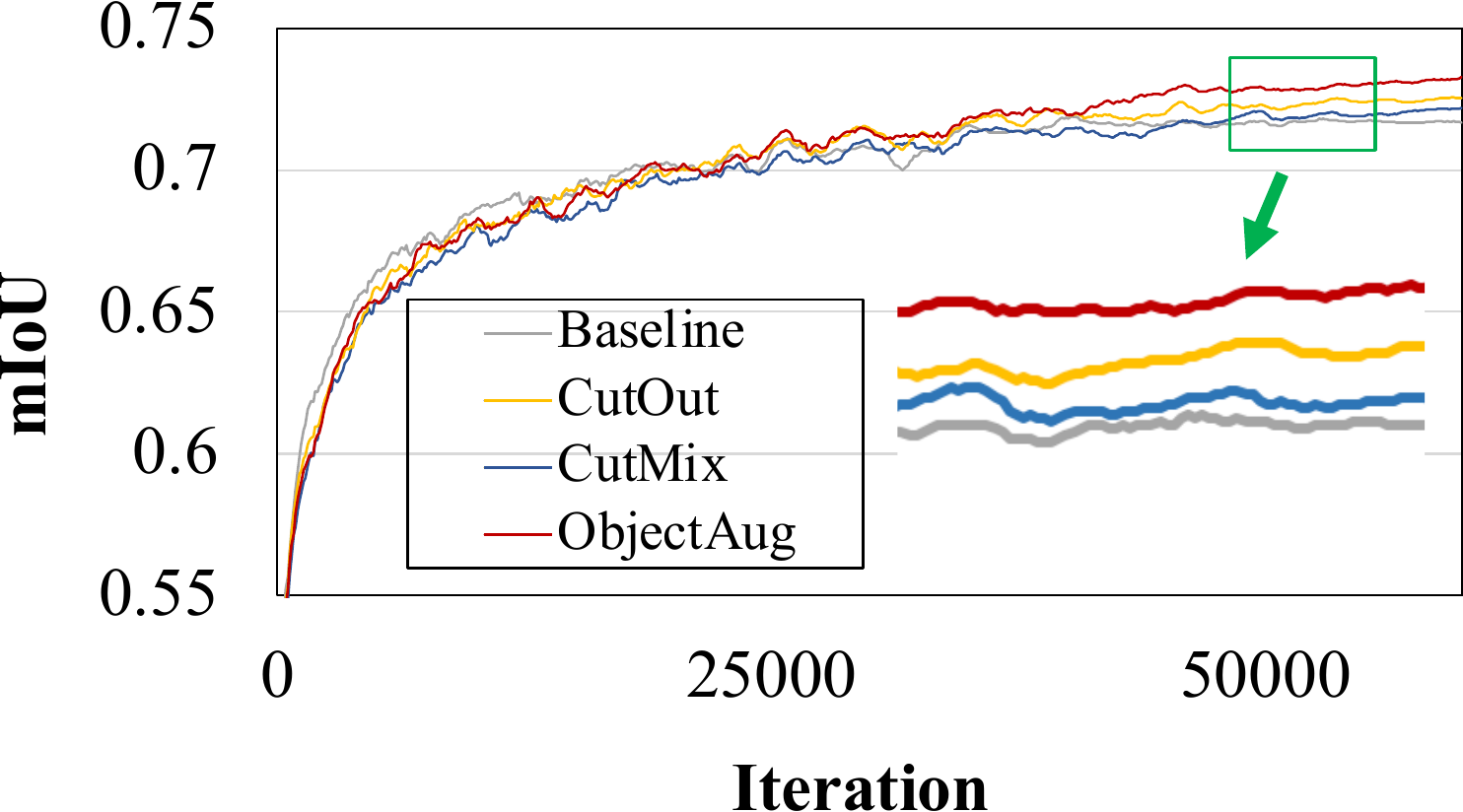}
		\caption{
			Comparison of mIoU curves over iteration times between CutOut, CutMix, and ObjectAug.
		}
		\label{fig-compact}
	\end{figure}

\begin{figure}[t]
		\centering
		\includegraphics[width=1\linewidth]{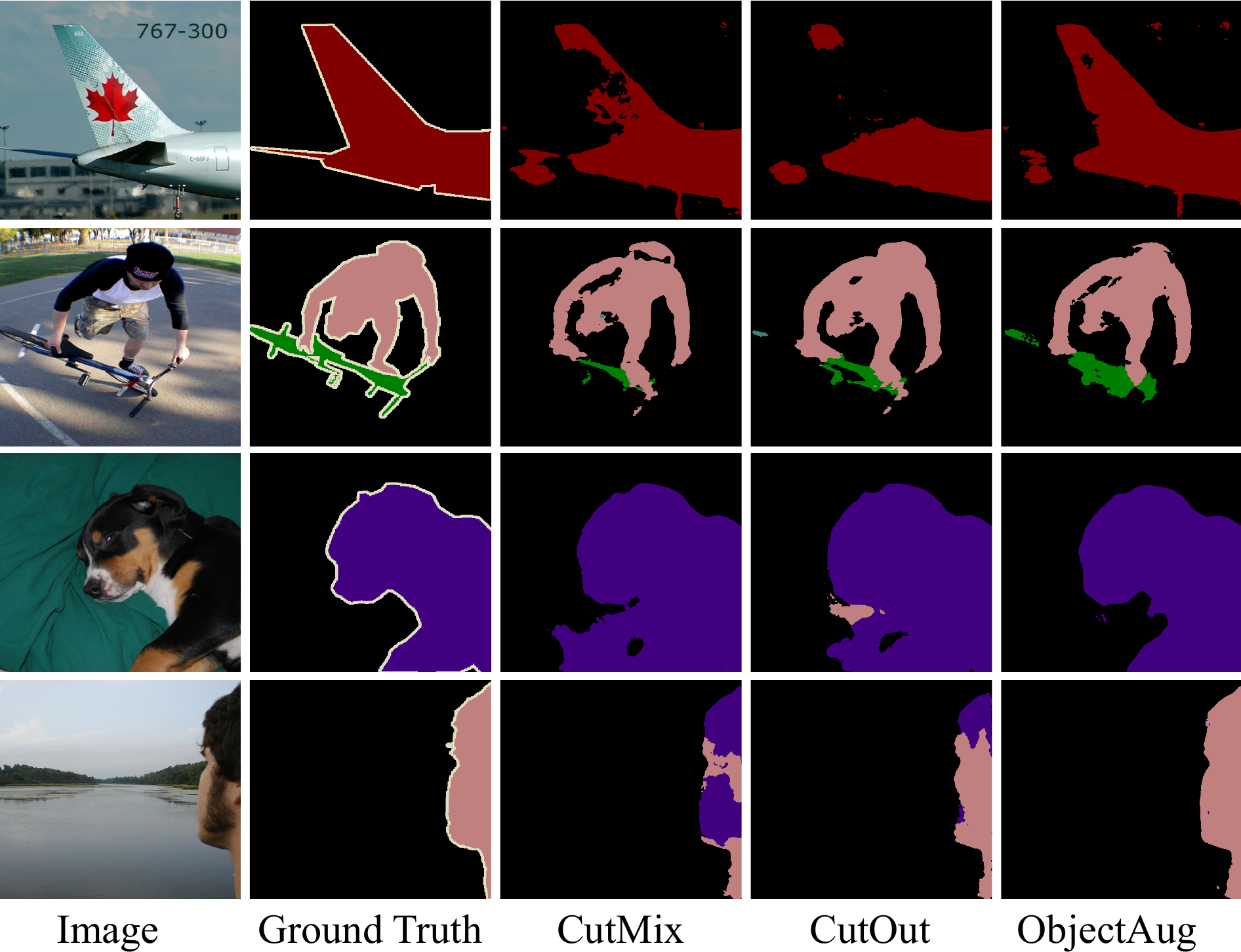}
		\caption{
			Qualitative comparison of ObjectAug and DNN-based data augmentation methods (CutOut, CutMix) on PASCAL VOC 2012.
		}
		\label{fig-display}
	\end{figure}

	We compared our ObjectAug with the current widely-used regularization technique, CutOut \cite{cutout} and CutMix \cite{cutmix}.
	%
	%
	%
	%
	Table \ref{tab-compact-pascal} details the comparison of our ObjectAug, CutOut, and CutMix.
	ObjectAug outperforms CutOut by 1.5\%, and CutMix by 1.1\%, respectively.
	ObjectAug achieves 73.8\% mIoU on PASCAL VOC 2012, which is 2.5\% higher than the baseline of 71.4\%. 
	Particularly, the state-of-the-art performance of 74.1\% is achieved by combining ObjectAug and CutOut.
	Figure \ref{fig-compact} shows the mIoU curve of them during training, and we can discover that the improvement of ObjectAug is rather stable.
	Qualitative comparison are shown in Figure \ref{fig-display}.
	Compared with CutOut and CutMix, ObjectAug obtains better performance especially on the boundaries.

	\begin{table}[ht]
		\centering
		\caption{
			Performance comparison of ObjectAug and DNN-based data augmentation methods on PASCAL VOC 2012.
			We use MobileNet based DeepLab V3plus model as our baseline. 
		}
		\begin{tabular}{l||c}
			\toprule[2pt]
			Method & mIoU (\%) \\ \hline	
			
			Baseline                        &  71.4        \\ \hline
			+ CutOut (16$\times$16, p = 0.5)   & 71.9    \\ 
			+ CutOut (16$\times$16, p = 1)      & 72.3    \\
			+ CutMix (p = 0.5)                         &  72.7   \\
			+ CutMix (p = 1)                           &  72.4    \\ 
			
			+ ObjectAug                 &\textbf{73.8}      \\ \hline
			
			+  CutOut (16$\times$16, p=0.5)  + ObjectAug   &  73.9     \\ 
			+  CutMix (p=0.5)  + ObjectAug        & \textbf{74.1}      \\ 
			\bottomrule[2pt]
			
		\end{tabular}
		\label{tab-compact-pascal}
	\end{table}

	\begin{table}[!t]
		\centering
		\caption{
			Ablation results of image inpainting. 
		}
		\begin{tabular}{l||c||c}
			\toprule[2pt]
			Method & Speed (s/iter) & mIoU (\%) \\ \hline	
			
			Baseline            & 0.698  &  71.4         \\ \hline
			+ ObjectAug(w/o fill)              &0.721 &  73.1  \\ 
			+ ObjectAug(random noises)      &0.729 &  73.3  \\
			+ ObjectAug(inpainting)    &0.827    & 73.8  \\
			
			\bottomrule[2pt]
			
		\end{tabular}
		\label{tab-inpaint}
	\end{table}

	\subsubsection{Ablation Analysis of Image Inpainting}
	
	CutOut and CutMix take different operations on the pixel artifacts.
	In this part, the ablation analysis of the image inpainting module is discussed. 
	%
	%
	%
	%
	As shown in Table \ref{tab-inpaint}, we can see that even if there is no processing on the pixel artifacts, our ObjectAug can still obtain a significant improvement by 1.7\%.
	In fact, the pixel artifacts brought by this method can be treated as a unique form of CutOut.
	Random noise filling also enhances the robustness to some extent.
	Meanwhile, the inpainting method achieves the best performance among the three processing methods.
	On the other hand, we also evaluated the training speed of the three methods.
	We can see that although ObjectAug using the inpainting method can achieve the best performance among the three methods, the speed is 18.5\% slower than the baseline.



	

	\begin{table}[t]
		\centering
		\caption{
			Result of impact discussion of category-aware coefficient. 
		}
		\begin{tabular}{l||cc||cc||c}
			\toprule[2pt]
			
			Method &H & NH & R&NR & Total \\ \hline	
			Baseline  &  57.2   &  84.3  &  70.1     &  72.6  & 71.4 \\ \hline
			Rarity-driven  &59.3  &85.8  &70.9 &75.3 & 73.2   \\ 
			Hard-driven    &60.2  &86.2  &70.6 &76.7 & 73.8    \\
			\bottomrule[2pt]
		\end{tabular}
		\label{tab-coefficient}
	\end{table}

	\subsubsection{Impact of Category-aware Strategy}
	%
	%
	%
	%
	%
	%
	In this section, we discuss two category-aware strategies: the rarity-driven coefficient and the hard-driven coefficient.
	We introduce two divisions for the 20 categories in PASCAL VOC 2012.
	%
	In the first division, the first ten (with small number of images) are divided into rare categories group (R), while the last ten (with large number of images) are divided into non-rare categories group (NR) to analyze the effect of rarity-driven coefficient.
	Similarly, the hard categories (H) and the non-hard categories (NH) are obtained to evaluate hard-driven coefficient in the second division.
	%
	%
	The comparison of two strategies is shown
	in Table \ref{tab-coefficient}.  
	Experimental results show that the segmentation performance of both categories is improved by 1.8-2.4\%, and the method shows more effectiveness on the hard-based rank (0.8\% higher).
	This is due to the fact that hard cases are more critical to effectively train DNNs.
	%
	
	\begin{figure}[htb]
		\centering
		\includegraphics[width=1\linewidth]{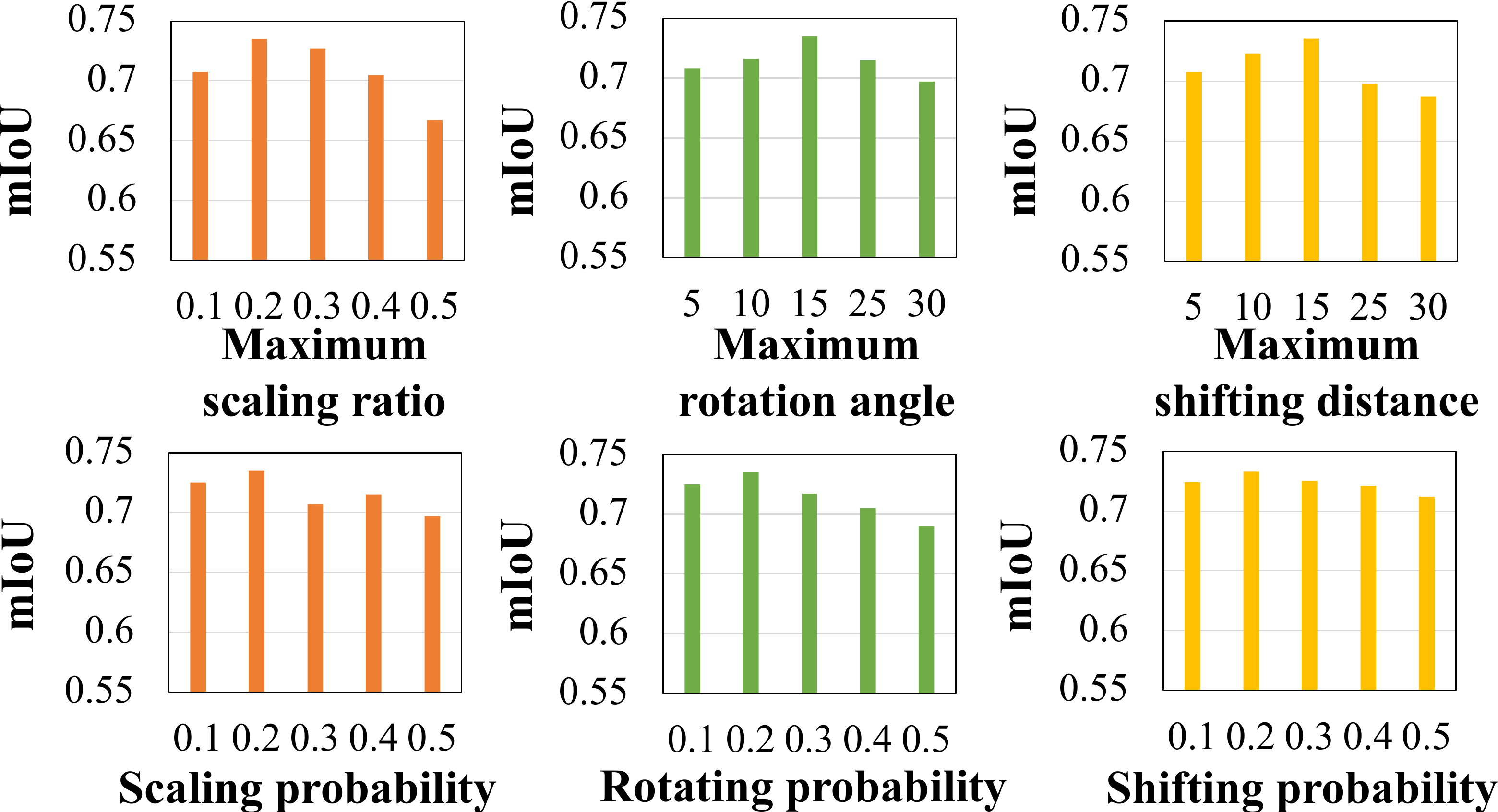}
		\caption{
			Result of impact discussion of hyper-parameters including random scaling, random rotation and random shift.
		}
		\label{fig-hyper}
	\end{figure}

	\subsubsection{Impact of Hyper-parameters}
	To study the impact of hyper-parameters, we need to assign two groups of hyper-parameters, configurations of each augmentation method and their possibilities.
	The first group includes the maximum scaling ratio $M_z$ in random scaling, maximum rotation angle $M_s$ in random rotation, and maximum shifting distance $M_s$ in random shifting.
	The second is their corresponding probabilities $P_z, P_r, P_s$. 
	To demonstrate the impact of these hyper-parameters on the performance, hyper-parameter settings are discussed on DeepLab V3plus (Mobile-Net).
	%
	We set $M_z = 1.2, M_r=\ang{15}, M_s=10$ as the base setting.
	%
	%
	Results are shown in Figure \ref{fig-hyper}.
	Notably, $M_z$, $M_r$, and $M_z$
	achieve the optimal performance with $M_z=0.2$, $M_r=15$, and $M_s=5$.
	Meanwhile, their probabilities of $[P_z, P_r, P_s]$ demonstrate a better performance with $[0.2,0.2,0.1]$.
	%
	%
	%

	\begin{table}[t]
		\centering
		\caption{
			Performance comparison of models with and without ObjectAug on Cityscapes.
		}
		\begin{tabular}{p{2.2cm}<{\centering}||p{1.5cm}<{\centering}||p{1.4cm}<{\centering}||p{1.5cm}<{\centering}}
			\toprule[2pt]
			Method      &Model &ObjectAug & mIoU (\%) \\  \hline

			\multirow{6}*{DeepLab V3plus}&\multirow{2}*{MobileNet }       & $\times$ & 72.0\\
			&        & \checkmark &  73.5 \\ \cline{2-4}
			
			
			&\multirow{2}*{ResNet-50 }      & $\times$ & 75.6  \\
			&     & \checkmark & 76.9  \\ \cline{2-4}
			
			&\multirow{2}*{ResNet-101 }      & $\times$ &77.4  \\
			&      & \checkmark & 78.5 \\ 
			\bottomrule[2pt]
		\end{tabular}
		\label{tab-cityscapes}
	\end{table}
	

	\begin{table}[t]
		\centering
		\caption{
			Performance comparison of ObjectAug and existing data augmentation methods on Cityscapes.
			%
		}
		\begin{tabular}{p{4.7cm}<{}||p{1.9cm}<{\centering}}
			\toprule[2pt]
			Method & mIoU (\%) \\ \hline	
			
			Baseline                        &  72.0        \\ \hline
			+ CutOut (16$\times$16, p = 1)      & 72.8    \\
			+ CutMix (p = 1)                           &  72.6    \\ 
			
			+ ObjectAug                 &\textbf{73.5}      \\ 
			
			\bottomrule[2pt]
			
		\end{tabular}
		\label{tab-compact-cityscapes}
	\end{table}
	
	\subsection{Extended Results on Cityscapes and CRAG}
	We further evaluated the effectiveness of ObjectAug on hard-to-segment data on the Cityscapes dataset, and its generalization on the CRAG dataset.
	\subsubsection{Results on Cityscapes}

	%
	%
	%
	%
	In Cityscapes, we consider the hard-to-segment objects as human, vehicle, object, and construction categories, which are augmented using ObjectAug.
	%
	As shown in Table \ref{tab-cityscapes}, our ObjectAug improves MobileNet, ResNet-50 and ResNet-101 based DeepLab V3plus by 1.5\%, 1.3\% and 0.9\% respectively.
	Table \ref{tab-compact-cityscapes} shows that ObjectAug also outperforms CutOut and CutMix by 0.7-0.9\%.
	Compared with CutOut and CutMix with an improvement of 0.8\% and 0.6\% respectively, the improvement of ObjectAug almost doubles.

	\begin{table}[h]
		\centering
		\caption{
			Performance comparison of models with and without ObjectAug on CRAG.
		}
		\begin{tabular}{p{2.7cm}<{\centering}||p{1.9cm}<{\centering}||p{1.9cm}<{\centering}}
			\toprule[2pt]
			Method &ObjectAug & mIoU (\%) \\  \hline		
			
			
			\multirow{2}*{FCN }       & $\times$ & 82.2 \\
			& \checkmark & 85.5 \\ \hline

			\multirow{2}*{U-Net }      & $\times$ &84.6 \\
			& \checkmark &87.2 \\ \hline
			
			\multirow{2}*{PSPNet }      & $\times$ &86.3 \\
			& \checkmark &89.1 \\ \hline
			
			\multirow{2}*{DeepLab V3}      & $\times$ &87.1 \\
			& \checkmark &89.7 \\ 
			
			\bottomrule[2pt]
		\end{tabular}
		\label{tab-crag}
	\end{table}

	\begin{table}[h]
		\centering
		\caption{
			Performance comparison of ObjectAug and existing data augmentation methods on CRAG.
			%
		}
		\begin{tabular}{p{4.7cm}<{}||p{1.9cm}<{\centering}}
			\toprule[2pt]
			Method & mIoU (\%) \\ \hline	
			
			Baseline                        &  84.6        \\ \hline
			+ CutOut (16$\times$16, p = 1)      & 85.5    \\
			+ CutMix (p = 1)                           &  85.3    \\ 
			
			+ ObjectAug                 &\textbf{86.2}      \\
			
			\bottomrule[2pt]
			
		\end{tabular}
		\label{tab-compact-crag}
	\end{table}

	\subsubsection{Results on CRAG}
	%
	Table \ref{tab-crag} shows the segmentation performance of FCN, U-Net \cite{unet}, PSPNet \cite{pspnet}, and DeepLab V3 \cite{deeplabv3} with and without ObjectAug on the CRAG dataset.
	We can see that ObjectAug can evidently improve the segmentation performance by 2.6-3.3\%.
	As shown in Table \ref{tab-compact-crag}, compared with other methods, ObjectAug can further improve the performance by 0.7-0.9\%.
	Compared with CutOut and CutMix with an improvement of 0.9\% and 0.7\% respectively, the improvement of ObjectAug almost doubles.


	\section{Conclusion}
	
	In this paper, we proposed a data augmentation method ObjectAug for semantic image segmentation.
	The proposed ObjectAug works at object level, which is completely different as the existing methods operating at image level.
	In addition, ObjectAug can support category-aware augmentation, and can be easily combined with the existing image-level augmentation methods to further enhance performance.
	The comprehensive experiments are conducted on both natural image and medical image datasets.
	Experiments cross various models and datasets demonstrate that our ObjectAug outperforms other augmentation methods and improve the segmentation performance.

	\section{Acknowledgment}
	
	This work was supported by the National Natural Science Foundation of China (No.62006050).
	

	{\small
		\bibliographystyle{IEEEtran}
		\bibliography{egbib}
	}

\end{document}